\documentclass[a4paper,oneside,11pt]{article} 

\usepackage[latin1]{inputenc}
\usepackage[T1]{fontenc}     
\usepackage{amsmath, amsthm} 
\usepackage{amsfonts,amssymb}

\usepackage{graphicx}        
\usepackage{subfig}          

\usepackage{array}           
\usepackage{multirow}        

\usepackage{color}           
\usepackage{bbm}             
\usepackage{dsfont}          
\usepackage{natbib}
\usepackage[section] {placeins}
\usepackage{verbatim}
\usepackage{hyperref}
\hypersetup{colorlinks, linkcolor=black, urlcolor=black} 

\usepackage{fullpage}

\newcommand{\OMIT}[1]{}
\newcommand{\cbr}[1]{\left\{#1\right\}}

\newcommand{\RR}{\mathbb{R}}

\newcommand{\Ecal}{\mathcal{E}}
\newcommand{\Gcal}{\mathcal{G}}
\newcommand{\Tcal}{\mathcal{T}}
\newcommand{\Rcal}{\mathcal{R}}

\newcommand{\ecoli}{\emph{E. coli}}
\newcommand{\sce}{\emph{S. cerevisiae}}

\begin{document}

\title{TIGRESS: Trustful Inference of Gene REgulation using Stability Selection.}

\author{Anne-Claire Haury$^{1,2,3,}$\footnote{To whom correspondence should be addressed.} , Fantine Mordelet$^{4}$, \\Paola Vera-Licona$^{1,2,3}$ and Jean-Philippe Vert$^{1,2,3,*}$\\ \\
$^{1}$ Centre for Computational Biology, Mines ParisTech, Fontainebleau, F-77300 France\\
$^{2}$ Institut Curie, Paris, F-75248, France\\
$^{3}$ U900, INSERM, Paris, F-75248, France\\
$^{4}$ Department of Computer Science, Duke University, Durham, NC 27708, USA.
}
\date{April 2012}
\OMIT{
\author{
\small\bf Anne-Claire Haury\thanks{To whom correspondence should
    be addressed: \small 35, rue Saint Honor\'e, F-77300 Fontainebleau, France.}\\
\small Mines ParisTech, CBIO\\
\small Institut Curie, Paris, F-75248\\
\small INSERM, U900, Paris, F-75248\\
\small \texttt{anne-claire.haury@mines-paristech.fr}
\and
\small\bf Fantine Mordelet\\
\small Mines ParisTech, CBIO\\
\small Institut Curie, Paris, F-75248\\
\small INSERM, U900, Paris, F-75248\\
\small \texttt{fantine.mordelet@mines-paristech.fr} \\
\and
\small\bf Paola Vera-Licona\\
\small Mines ParisTech, CBIO\\
\small Institut Curie, Paris, F-75248\\
\small INSERM, U900, Paris, F-75248\\
\small \texttt{paola.vera-licona@curie.fr} \\
\and
\small\bf Jean-Philippe Vert \\
\small Mines ParisTech, CBIO\\
\small Institut Curie, Paris, F-75248\\
\small INSERM, U900, Paris, F-75248\\
\small \texttt{jean-philippe.vert@mines-paristech.fr} \\
}
}
\maketitle

\begin{abstract}

{\bf Background}. Inferring the structure of gene regulatory networks (GRN) from a collection of gene expression data has many potential applications, from the elucidation of complex biological processes to the identification of potential drug targets. It is however a notoriously difficult problem, for which the many existing methods reach limited accuracy.

{\bf Results}. In this paper, we formulate GRN inference as a sparse regression problem and investigate the performance of a popular feature selection method, least angle regression (LARS) combined with stability selection, for that purpose. We introduce a novel, robust and accurate scoring technique for stability selection, which improves the performance of feature selection with LARS. The resulting method, which we call TIGRESS (for Trustful Inference of Gene REgulation with Stability Selection), was ranked among the top GRN inference methods in the DREAM5 gene network reconstruction challenge. We investigate in depth the influence of the various parameters of the method, and show that a fine parameter tuning can lead to significant improvements and state-of-the-art performance for GRN inference. 

{\bf Conclusions}. TIGRESS reaches state-of-the-art performance on benchmark data. This study confirms the potential of feature selection techniques for GRN inference.   

{\bf Availability}. Code and data are available on http://cbio.ensmp.fr/$\sim$ahaury. Moreover, running TIGRESS online is possible on GenePattern: \url{http://www.broadinstitute.org/cancer/software/genepattern/}.

\end{abstract}
\section{Background}
In order to meet their  needs and adapt to changing environments, cells have developed various mechanisms to regulate the production of the thousands of proteins they can synthesize. Among them, the regulation of gene expression by transcription factors (TF) is preponderant: by binding to the promoter regions of their target genes (TG), TF can activate or inhibit their expression. Deciphering and understanding TF-TG regulations has many potential far-reaching applications in biology and medicine, ranging from the \emph{in silico} modelling and simulation of the gene regulatory network (GRN) to the identification of new potential drug targets. However, while many TF-TG regulations have been experimentally characterized in model organisms, the systematic experimental characterization of the full GRN remains a daunting task due to the large number of potential regulations. 

The development of high-throughput methods, in particular DNA microarrays, to monitor gene expression on a genome-wide scale has promoted new strategies to elucidate GRN. By systematically assessing how gene expression varies in different experimental conditions, one can try to \emph{reverse engineer} the TF-TG interactions responsible for the observed variations. Not surprisingly, many different approaches have been proposed in the last decade to solve this GRN reverse engineering problem from collections of gene expression data. When expression data are collected over time, for example, several methods have been proposed 
to construct dynamic models where TF-TG interactions dictate how the expression level of a TG at a given time allows to predict the expression levels of its TG in subsequent times~\citep{Arkin1997test,Liang1998Reveal,Chen1999Modeling,Akutsu2000Algorithms,Yeung2002Reverse,Tegner2003Reverse,Gardner2003Inferring,Chen2005stochastic,Bernardo2005Chemogenomic,Bansal2006Inference}. When expression data are not limited to time series, many methods attempt to capture statistical association between the expression levels of TG and candidate TF using correlation or information-theoretic measures or mutual information \citep{Butte2000Discovering,Margolin2006ARACNE,Faith2007Large-scale} or take explicit advantage of perturbations in the experiments such as gene knock-downs \citep{Rice2005Reconstructing}. The difficulty to separate direct from indirect regulations has been addressed with the formalism of Bayesian networks \citep{Friedman2000Using,Hartemink2002Using,Friedman2004Inferring}, or by formulating the GRN inference problem as a feature selection problem \cite{Huynh-Thu2010Inferring}. We refer to \cite{Markowetz2007Inferring,Marbach2010Revealing} for detailed reviews and comparisons of existing methods. 

Recent benchmarks and challenges have highlighted the good performance of methods which formalize the GRN inference problem as a regression and feature selection problem, namely, 
identifying a small set of TF whose expression levels are sufficient to predict the expression level of each TG of interest \citep{Meinshausen2006High}. This idea underlies the Bayesian network formalism \citep{Friedman2000Using}, but is more directly addressed by GENIE3 \cite{Huynh-Thu2010Inferring}, a method which uses random forests to identify TF whose expression levels are predictive for the expression level of each TG, and which is now recognized as state-of-the-art on several benchmarks \citep{Huynh-Thu2010Inferring,Marbach2010Revealing}. Feature selection with random forests remains however poorly understood theoretically, and one may wonder how other well-established statistical and machine learning techniques for feature selection would behave to solve the GRN inference problem.

In this paper, we investigate the performance of a popular feature selection method, least angle regression (LARS) \citep{Efron2004Least} combined with stability selection \citep{Bach2008Bolasso,Meinshausen2010Stability}, for GRN inference. LARS is a computationally efficient procedure for multivariate feature selection, closely related to Lasso regression \citep{Tibshirani1996Regression}. 
We introduce a novel, robust and accurate scoring technique for stability selection, which improves the performance of feature selection with LARS. The resulting method, which we call TIGRESS (for Trustful Inference of Gene REgulation with Stability Selection), was ranked among the top GRN inference methods in the DREAM5 gene reconstruction challenge \citep{Marbach2012Wisdom}. We furthermore investigate in depth the influence of the various parameters of the method, and show that a fine parameter tuning can lead to significant improvements and state-of-the-art performance for GRN inference. Overall this study confirms the potential of state-of-the-art feature selection techniques for GRN inference.

\section{Methods}

\subsection{Problem formulation}
We consider a set of $p$ genes $\Gcal = [1,p]$, including a subset $\Tcal\subset[1,p]$ of transcription factors, among which we wish to discover direct regulations of the form $(t,g)$ for $t\in\Tcal$ and $g\in\Gcal$. We do not try to infer self-regulation, meaning that for each target gene $g\in\Gcal$ we define the set of possible regulators as $\Tcal_g = \Tcal \backslash \{g\}$ if $g\in\Tcal$ is itself a transcription factor, and $\Tcal_g = \Tcal$ otherwise. The set of all candidate regulations is therefore $\Ecal = \cbr{(t,g) , g\in\Gcal , t\in\Tcal_g}$, and the GRN inference problem is to identify a subset of true regulations among $\Ecal$.

For that purpose, we assume we have gene expression measurements for all genes $\Gcal$ in $n$ experimental conditions. Although the nature of the experiments may vary and typically include
knock-down or knock-out experiments and even replicates, for simplicity we do not exploit this information and only consider the $n\times p$ data matrix of expression levels $X$ as input for the GRN inference problem. Each row of $X$ corresponds to an experiment and each column to a gene. We assume that the expression data have been pre-processed for quality control and missing values imputation.

In order to infer the regulatory network \OMIT{$\Rcal$} from the expression data $X$, we compute a score $s:\Ecal\rightarrow\RR$ to assess the evidence that each candidate regulation is true, and then predict as true regulation the pairs $(t,g)\in\Ecal$ for which the evidence $s(t,g)$ is larger than a threshold $\delta$. We let $\delta$ as a user-controlled parameter,
where larger $\delta$ values correspond to less predicted regulations,  and only focus on designing a significance score $s(t,g)$ that leads to "good" prediction for some values of $\delta$. In other words, we only focus on finding a good ranking of the candidate regulations $\Ecal$, by decreasing score, such that true regulations tend to be at the top of the list; we let the user control the level of false positive and false negative predictions he can accept.

\subsection{GRN inference with feature selection methods}
Many popular methods for GRN inference are based on such a score. For example, the correlation or mutual information between the expression levels of $t$ and $g$ along the different experiments is a popular way to score candidate regulations \citep{Butte2000Discovering,Margolin2006ARACNE,Faith2007Large-scale}. A drawback of such direct approaches is that it is then difficult to separate direct from indirect regulations. For example, if $t_1$ regulates $t_2$ which itself regulates $g$, then the correlation or mutual information between $t_1$ and $g$ is likely to be large, although $(t_1,g)$ is not a direct regulation. Similarly, if $t_1$ regulates both $t_2$ and $g$, then $t_2$ and $g$ will probably be very correlated, even if there is no direct regulation between them. In order to overcome this problem, a possible strategy is to post-process the predicted regulations and try to remove regulations likely to be indirect because they are already explained by other regulations \citep{Margolin2006ARACNE}. Another strategy is, given a target gene $g\in\Gcal$, to \emph{jointly} estimate the scores $s(t,g)$ for all candidate regulators $t \in \Tcal_g$ simultaneously, with a method able to capture the fact that a large score for a candidate regulation $(t,g)$ is not needed if the apparent correlation between $t$ and $g$ is already explained by other, more likely regulations. 

Mathematically, the latter strategy is closely related to the problem of \emph{feature selection} in statistics, as already observed and exploited by several authors \cite{Meinshausen2006High,Huynh-Thu2010Inferring}. More specifically, for each target gene $g\in\Gcal$, we consider the regression problem where we wish to predict the expression level of $g$ from the expression level of its candidate regulators $t\in\Tcal_g$:
\begin{equation}\label{eq:reg}
X_g = f_g(X_{\Tcal_g}) + \epsilon\,,
\end{equation}
where $X_i$ represents the expression level of the $i$-th gene across different experiments (modelled as a random variable), $X_{\Tcal_g}=\cbr{X_t\,,\,t\in\Tcal_g}$ is the set of expression levels of the candidate transcription factors for gene $g$, and $\epsilon$ is some noise. Any linear or nonlinear statistical method for regression can potentially be used to infer $f_g$ from the observed expression data.  However, we are not directly interested in the regression function $f_g$, but instead in the identification of a small set of transcription factors which are sufficient to provide a good model for $X_g$. We therefore need a score $s_g(t)$ for each candidate transcription factor $t\in\Tcal_g$ to assess how likely it is to be involved in the regression model $f_g$. For example, if we model $f_g$ as a linear function
\begin{equation}\label{eq:linreg}
f_g(X_{\Tcal_g}) = \sum_{t\in\Tcal_g} \beta_{t,g} X_t\,,
\end{equation}
then the score $s_g(t)$ should typically assess the probability that $\beta_{t,g}$ is non-zero \citep{Meinshausen2006High}. More general models are possible, for example \cite{Huynh-Thu2010Inferring} model $f_g$ with a random forest \cite{Breiman2001Random} and score a predictor $s_g(t)$ with a variable importance measure specific to this model. Once a score $s_g(t)$ is chosen to assess the significance of each transcription factor in the target-gene-specific regression model (\ref{eq:reg}), we can combine them across all target genes by defining the score of a candidate regulation $(t,g)\in\Ecal$ as $s(t,g) = s_g(t)$, and rank all candidate regulations by decreasing score for GRN inference.

\subsection{Feature selection with LARS and stability selection} \label{sec:lasso}

We now propose a new scoring function $s_g(t)$ to assess the significance of a transcription factor $t\in\Tcal_g$ in the regression model (\ref{eq:reg}). Our starting point to define the scoring function is the LARS method for feature selection in regression \citep{Efron2004Least}. LARS models the regression function (\ref{eq:reg}) linearly, {\textit i.e.} it models the expression of a target gene as a linear combination of the expression of its transcription factors, as in (\ref{eq:linreg}). Starting from a constant model where no TF is used, it iteratively adds TF in the model to refine the prediction of $X_g$. Contrary to classical forward stepwise feature selection \citep{Weisberg1981Applied,Hastie2001elements}, LARS does not fully re-optimize the fitted model when a new TF is added to the model, but only refines it partially. This results in a statistically sound procedure for feature selection, akin to forward stage-wise linear regression and the Lasso \cite{Tibshirani1996Regression,Hastie2001elements}, and a very efficient computational procedure. In practice, after $L$ steps of the LARS iteration, we obtain a ranked list of $L$ TF selected for their ability to predict the expression of the target gene of interest. Efficient implementations of LARS exist in various programming languages including R (\verb=lars= package, \citep{Efron2004Least}) and MATLAB (\verb=SPAM= toolbox, \citep{Mairal2010Online}). Since the selection of TF is iterative, LARS has the potential to disregard indirect regulations.

The direct use of LARS to score candidate regulations has, however, two shortcomings. First, LARS can be very sensitive and unstable in terms of selected features when there exist high correlations between different explanatory variables. Second, it only provides a ranking of the TF, for each TG of interest, but does not provide a score $s_g(t)$ to quantify the evidence that a TF $t$ regulates a target gene $g$. Since we want to aggregate the predicted regulations across all target genes to obtain a global ranking of all candidate regulations, we need such a score.

\begin{figure}[!ht]
  \centering
  \includegraphics[width=.5\textwidth]{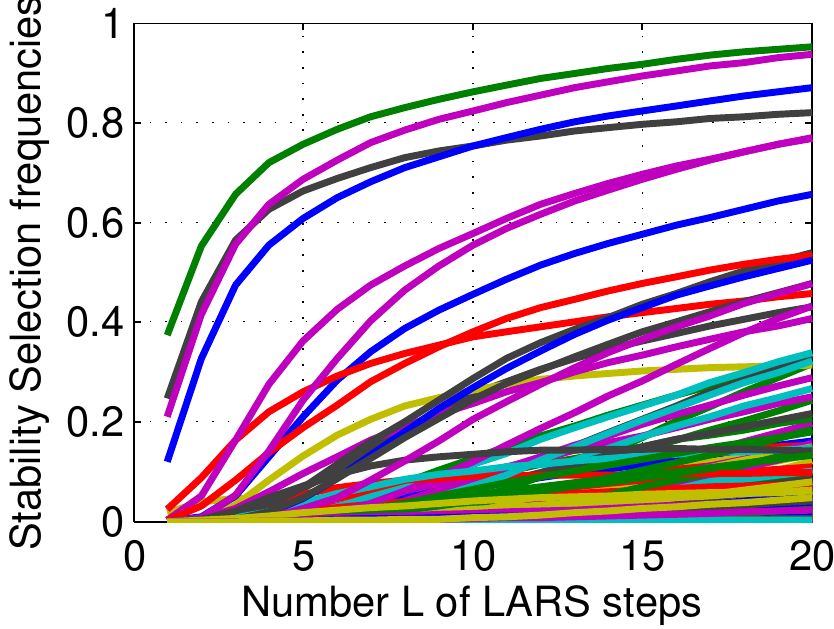}
\caption{Illustration of the stability selection frequency $F(g,t,l)$ for a fixed target gene $g$. Each curve represents a TF $t\in\Tcal_g$, and the horizontal axis represents the number $l$ of LARS steps. $F(g,t,l)$ is the frequency with which $t$ is selected in the first $l$ LARS steps to predict $g$, when the expression matrix is randomly perturbed by selecting only a limited number of experiments and randomly weighting each expression array. For example, the TF corresponding to the highest curve was selected $57\%$ of the time at the first LARS step, and $81\%$ of the time in the first two LARS steps.}
 \label{fig:ss}
\end{figure}

To overcome both issues, we do not directly score candidate regulations with the LARS, but instead perform a procedure known as \emph{stability selection} \cite{Meinshausen2010Stability} on top of LARS. The general idea of stability selection is to run a feature selection method many times on randomly perturbed data, and score each feature by the number of times it was selected. It was shown that stability selection can reduce the sensitivity of LARS and Lasso to correlated features, and improve their ability to select correct features \cite{Bach2008Bolasso,Meinshausen2010Stability}. In addition, it provides a score for each feature, which can then be aggregated over different regression problems, \emph{i.e.} different target genes in our case. More precisely, to score the candidate target genes $t\in\Tcal_g$ of a given target gene $g$ using LARS with stability selection, we fix a (large) number of iterations $R$, and repeat $R/2$ times the following iterations: we randomly split the experiments into two halves of equal or approximately equal size, we multiply the expression levels of the candidate transcription factors in $\Tcal_g$ on each microarray by a random number uniformly sampled on the interval $[\alpha,1]$ for some $0\leq\alpha\leq 1$, and we run the LARS method for $L>0$ steps on the two resulting reduced and reweighed expression matrices. We therefore perform a total of $R$ LARS runs on randomly modified expression matrices. For each run, the result of LARS after $L$ steps is a ranked list of $L$ TF. After the $R$ runs, we record for each $g\in\Gcal$, $t\in\Tcal_g$ and $l\in[1,L]$ the frequency $F(g,t,l)$ with which the TF $t$ was selected by the LARS in the top $l$ features to predict the expression of gene $g$. We divide the frequency by $R$ to obtain a final score between $0$ and $1$, $1$ meaning that $t$ is always selected by LARS in the top $l$ features to predict the expression level of $g$, and $0$ that is is never selected. Figure~\ref{fig:ss} displays graphically these frequencies, for a given gene $g$ fixed, all candidate TF in $\Tcal_g$, and $l=1,\ldots, 15$. When $l$ increases, the frequency $F(g,t,l)$ for fixed $g$ and $t$ is non-decreasing because the LARS method selects increasing sets of TF at each step. In addition, since the total number of TF selected after $l$ LARS steps is always equal to $l$, taking the average over the $R$ LARS runs leads to the equality $\sum_{t\in\Tcal_g} F(g,t,l) = l$, for any gene $g$ and LARS step $l$.

Once the frequency table $F(g,t,l)$ is computed for $l=1,\ldots,L$, we need to convert it into a unique score $s(t,g)$ for each candidate pair $(t,g)$. The original stability selection score \cite{Bach2008Bolasso,Meinshausen2010Stability} is simply defined as the frequency of selection in the top $L$ variables, {\textit i.e.},
\begin{equation}\label{eq:originalscore}
s_{original}(t,g) = F(g,t,L)\,.
\end{equation}
As suggested by Figure~\ref{fig:ss}, this score may be very sensitive to the choice of $L$. In particular, if $L$ is too small, many TF may have zero score (because there are never selected in the top $L$ TFs), but when $L$ is too large, several TF may have the same score $1$ because they are always selected in the top $L$ TFs. To alleviate this possible difficulty, we propose as an alternative score to measure the \emph{area under each curve} up to $L$ steps, \emph{i.e.} to consider the following \emph{area score}:
\begin{equation}\label{eq:areascore}
s_{area}(t,g) = \frac{1}{L}\sum_{l=1}^L F(g,t,l)\,.
\end{equation}
The difference between $s_{original}(t,g)$ and $s_{area}(t,g)$ becomes clear if we consider the rank of $t$ in the list produced by LARS in one run as a random variable $H_t$ (with $H_t=1$ meaning that $t$ is ranked first by LARS). $F(g,t,l)$ is then, by definition, the empirical probability $P(H_t\leq l)$ that $H_t$ is not larger than $l$.  The original score has therefore an obvious interpretation as $P(H_t \leq L)$, which we can rewrite as:
$$
s_{original}(t,g)  =  E\left[\phi_{original}(H_t)\right] \quad\text{with}\quad \phi_{original}(h) = 
\begin{cases}
1 & \text{if }h\leq L\,,\\
0&\text{otherwise.}
\end{cases}
$$
Interestingly a small computation shows that the area score has a similar probabilistic interpretation:
\begin{equation*}
\begin{split}
s_{area}(t,g) &= \sum_{l=1}^L F(g,t,l)\\
&= \sum_{l=1}^L P(H_t \leq l)\\
&=  \sum_{l=1}^L \sum_{h=1}^l P(H_t = h)\\
&= \sum_{h=1}^L (L+1-h) P(H_t = h)\\
&= E\left[\phi_{area}(H_t)\right] \,,
\end{split}
\end{equation*}
with
$$
 \phi_{area}(h) = 
\begin{cases}
L+1-h & \text{if }h\leq L\,,\\
0&\text{otherwise.}
\end{cases}
$$
In other words, both the original and the area scores can be expressed as $E\left[\phi(H_t)\right]$, although with a different function $\phi$. While the original score only assesses how often a feature ranks in the top $L$, the area score additionally takes into account the  value of the rank, with features more rewarded if they are not only in the top $L$ but also frequently with a small rank among the top $L$. Since $s_{area}$ integrates the frequency information over the full LARS path up to $L$ steps, it should be less sensitive than $s_{original}$ to the precise choice of $L$, and should allow to investigate larger values of $L$ without saturation effects when several curves hit the maximal frequency of $1$. We note that other scores of the form $E\left[\phi(H_t)\right] $ for non-increasing function $\phi$ could be investigated as well.

\subsection{Parameters of TIGRESS}\label{sec:parameters}
In summary, the full procedure for scoring all candidate edges in $\Ecal$, which we call TIGRESS, splits the GRN inference problem into $p$ independent regression problems taking each target gene $g\in\Gcal$ in turn, and scores each candidate regulation $(t,g)$ for a candidate TF $t\in \Tcal_g$ with the original (\ref{eq:originalscore}) or area (\ref{eq:areascore}) stability score applied to LARS feature selection. In addition to the choice of the scoring method (original or area), the parameters of TIGRESS are (i) the number of runs $R$ performed in stability selection to compute the scores, (ii) the number of LARS steps $L$, and (iii) the parameter $\alpha\in[0,1]$ which controls the random re-weighting of each expression array in each stability selection run. Apart from $R$ that should be taken as large as possible to ensure that frequencies are correctly estimated, and is only limited by the computational time we can afford to run TIGRESS, the influence of $\alpha$ and $L$ on the final performance of the method are not obvious. Taking $\alpha=1$ means that no weight randomization is performed on the different expression arrays, while $\alpha=0$ leads to maximal randomization. \cite{Meinshausen2010Stability} advocate that a value between $0.2$ and $0.8$ is often a good choice. Regarding the choice of $L$, \cite{Meinshausen2010Stability} mentions that it has usually little influence on the result, but as discussed above, the choice of a good range of values may not be trivial in particular for the original score. We investigate below in detail how the performance of TIGRESS depends on the scoring method and on these parameters $R$, $\alpha$ and $L$.

\subsection{Other GRN inference methods}

We experimentally compare TIGRESS to several other GRN inference methods. We use the MATLAB implementations of CLR \citep{Faith2007Large-scale} and GENIE3 \citep{Huynh-Thu2010Inferring}. We run ARACNE \citep{Margolin2006ARACNE} using the R package \verb=minet=. We keep default parameter values for each of these methods. Results borrowed from the DREAM5 challenge \cite{Marbach2012Wisdom} were directly obtained by each participating team.

\subsection{Performance evaluation}
Given a gene expression data matrix, each GRN inference method outputs a ranked list of putative TF-TG regulations. Taking only the top $K$ predictions in this list, we can compare them to known regulations to assess the number of true positives ($TP$, the number of known regulations in the top $K$ predictions), false positives ($FP$, the number of predicted regulations in the top $K$ which are not known regulations), false negatives ($FN$, the number of known interactions which are not in the top $K$ predictions) and true negatives ($TN$, the number of pairs not in the top $K$ predictions which are not known regulations). We then compute classical statistics to summarize these numbers for a given $K$, including precision ($TP/(TP+FP)$), recall ($TP/(TP+FN)$), and fall-out ($FP/(FP+TN)$). We assess globally how these statistics vary with $K$ by plotting the receiver operating characteristic (ROC) curve (recall as a function of fall-out) and the precision-recall curve (precision as a function of recall), and computing the area under these curves (respectively AUROC and AUPR) normalized between $0$ and $1$.

For the datasets of DREAM5, we further compute a $P$-value for the AUROC and AUPR scores, based on all DREAM5 participants' predictions. This method, which was used by the DREAM5 organizers to rank the teams, involves randomly drawing edges from the teams' prediction lists and computing the probabilities of obtaining an equal or larger AUPR (resp. AUROC) by chance. More precisely, random lists are constructed as follows: for each row of the predicted list, an edge at the same position is drawn at random from all predictions. For an ensemble of such random lists, the areas under the curves are computed, allowing to estimate a random distribution. $P$-values were obtained by extrapolating the resulting histogram. We refer to \cite{Marbach2012Wisdom} for more details on this scoring scheme. Finally, we compute the so-called \emph{overall score} for a GRN inference method by integrating the AUROC and AUPR $P$-values as follows:
\begin{equation}\label{overall_score}
\text{overall score}=\frac{1}{2} \ln(p_{AUPR}\,p_{AUROC})\,.
\end{equation}

\section{Data}
We evaluate the performance of TIGRESS and other GRN inference methods on four benchmark datasets, each consisting of a compendium of gene expression data, a list of known TF, and a gold standard set of verified TF-TG regulations which we ideally would like to recover from the expression data only. Expression data are either simulated or experimentally measured under a wide range of genetic, drug and environmental perturbations. Table~\ref{tab:data} summarizes the statistics of these four networks.

The first three benchmarks are taken from the DREAM5 challenge \cite{Marbach2012Wisdom}. Network 1 is a simulated dataset. Its topology and dynamics were modelled according to known GRN, and the expression data were simulated using the \emph{GeneNetWeaver} software \citep{Schaffter2011GeneNetWeaver}. We refer the interested reader to \cite{Marbach2009Generating,Marbach2010Revealing} for more information on this network. The second and third benchmarks are Network 3 and Network 4 of the DREAM5 competition, corresponding respectively to real expression data for \ecoli{} and \sce. Note that we do not use in our experiments Network 2 of DREAM5, because no verified TF-TG is provided for this dataset consisting in expression data for \emph{S. aureus}.

Additionally, we run experiments on the \ecoli{} dataset from \cite{Faith2007Large-scale}, which has been widely used as a benchmark in GRN inference literature. The expression data was downloaded from the Many Microbe Microarrays ($M^{3D}$) database \cite{Faith2008Many} (version $4$ build $6$). It consists in $907$ experiments and $4297$ genes. We obtained the gold standard data from RegulonDB \cite{Gama-Castro2011RegulonDB} (version 7.2, May 6th, 2011) that contains $3812$ verified interactions among $1525$ of the genes present in the microarrays experiments. 

As a pre-processing step, we simply mean-center and scale to unit variance the expression levels of each gene within each compendium.
\begin{table}[!ht]
\centering
\begin{tabular}{|c||c|c|c|c|}
\hline
\textbf{Network} 			&	 \textbf{$\sharp$ TF} 	& \textbf{$\sharp$  Genes} 	&	\textbf{$\sharp$ Chips} & \textbf{$\sharp$ Verified interactions}\\ \hline
DREAM5 Network 1 (in-silico) &	195   		&	1643 		&	805 & 4012\\ \hline
DREAM5 Network 3 (\ecoli)		&	334 			&	4511 		&	805 & 2066\\ \hline
DREAM5 Network 4 (\sce)		&	333 			&	5950 		&	536 & 3940 \\ \hline
\ecoli{} Network from \cite{Faith2007Large-scale} &    180   & 1525  & 907 & 3812 \\ \hline
\end{tabular}
\caption{Datasets used in our experiments.}
\label{tab:data}
\end{table}

\section{Results} \label{sec:results}

\subsection{DREAM5 challenge results}\label{sec:dream5}

In $2010$ we participated to the DREAM5 Network Inference Challenge, an open competition to assess the performance of GRN methods \citep{Marbach2012Wisdom}. Participants were asked to submit a ranked list of predicted interactions from four matrices of gene expression data. At the time of submission, no further information was available to participants (besides the list of TF), in particular the "true" network of verified interactions for each dataset was not given. After submissions were closed, the organizers of the challenge announced that one network (Network 1) was a simulated network with simulated expression data, while the other expression datasets were real expression data collected for \ecoli{} (Network 3) and \sce{} (Network 4), respectively. \OMIT{ Since no "true" GRN is available for \emph{S. aureus}, participants were only scored on their capacity to recover the gold standard GRN of Network 1, 3 and 4.} Teams were ranked for each network by decreasing overall score \eqref{overall_score}, and an overall ranking was proposed based on the average of the overall scores over the three networks.

We submitted predictions for all networks with a version of TIGRESS which we did not try to optimize, which we refer to as \emph{Naive TIGRESS} below. Naive TIGRESS is the variant of TIGRESS which scores candidate interactions with the original score (\ref{eq:originalscore}) and uses the arbitrarily fixed parameters $\alpha=0.2$, $L=5$, $R_1=4,000$, $R_3=R_4=1,000$, where $R_i$ refers to the number of runs for network $i$. The number of runs were simply set to ensure that TIGRESS would finish within 2 days on a single-core laptop computer. $R_1$ is larger than $R_3$ and $R_4$ because the size of network $1$ is smaller than that of networks $3$ and $4$, implying that each TIGRESS run is faster. The choice $\alpha=0.2$ followed previous suggestions for the use of stability selection \citep{Meinshausen2010Stability}, while the choice $L=5$ roughly corresponded to the largest value for which no TF-TG pair had a score of $1$.

Naive TIGRESS was among the top GRN prediction methods at DREAM5, ranking second among 29 participating teams in the \emph{in silico} network challenge, and third overall. Table \ref{tab:dream_results} summarizes the results of the first three teams in average overall score.
\begin{table}[!ht] 
\small
\centering
\begin{tabular}{|l|c|c|c|c|c|c|c|c|c|} 
\hline 
\textbf{Teams} & \multicolumn{3}{|c|}{\textbf{Network 1}} &   \multicolumn{3}{|c|}{\textbf{Network 3}} &  \multicolumn{3}{|c|}{\textbf{Network 4}}\\ \hline \hline
& \emph{AUPR} & \emph{AUROC}  & \emph{Score} &  \emph{AUPR} & \emph{AUROC} & \emph{Score} & \emph{AUPR} & \emph{AUROC} & \emph{Score} \\ \hline
GENIE3 \cite{Huynh-Thu2010Inferring}  &0.291	&0.815 & 104.65	&0.093	&0.617 & 14.79 &0.021&	0.518  & 1.39\\   \hline
ANOVA-based \cite{Kueffner2012Inferring} & 0.245	&0.780 & 	53.98 &0.119	&0.671 & 45.88 &0.022&	0.519 & 2.21 \\ \hline 
\textbf{Naive TIGRESS} & \textbf{0.301}&\textbf{0.782} & \textbf{87.80}&\textbf{0.069}	&\textbf{0.595}& \textbf{4.41}&\textbf{0.020}	&\textbf{0.517} & \textbf{1.08}\\ \hline
\end{tabular} 
\caption{AUPR, AUROC and minus the logarithm of related p-values for all DREAM5 Networks and the three best teams.}
\label{tab:dream_results}
\end{table} 
The winning method, both \emph{in silico} and overall, was the GENIE3 method of \cite{Huynh-Thu2010Inferring}. GENIE3 already won the DREAM4 challenge, confirming its overall state-of-the-art performance. It had particularly strong performance on the \emph{in silico} network, and more modest performance on both \emph{in vivo} networks. The ANOVA-based method of \citep{Kueffner2012Inferring} ranked second overall, with particularly strong performance on the \ecoli{} network. Naive TIGRESS ranked third overall, with particularly strong performance on the \emph{in silico} network, improving over GENIE3 in terms of AUPR.

Interestingly, GENIE3 and TIGRESS follow a similar formulation of GRN inference as a collection of feature selection problems for each target gene, and use similar randomization-based techniques to score the evidence of a candidate TF-TG regulation.  The main difference between the two methods is that GENIE3 aggregates the features selected by decision trees, while TIGRESS aggregates the features selected by LARS.  The overall good results obtained by both methods suggest that this formalism is particularly relevant for GRN inference.

\subsection{Influence of TIGRESS parameters}

In this section, we provide more details about the influence of the various parameters of TIGRESS on its performance, taking DREAM5 \emph{in silico} network as benchmark dataset. Obviously the improvements we report below would require confirmation on new datasets not used to optimize the parameters, but they shed light on the further potential of TIGRESS and similar regression-based method when parameters are precisely tuned.

\begin{figure}[!ht]
  \centering
  \includegraphics[width=\textwidth]{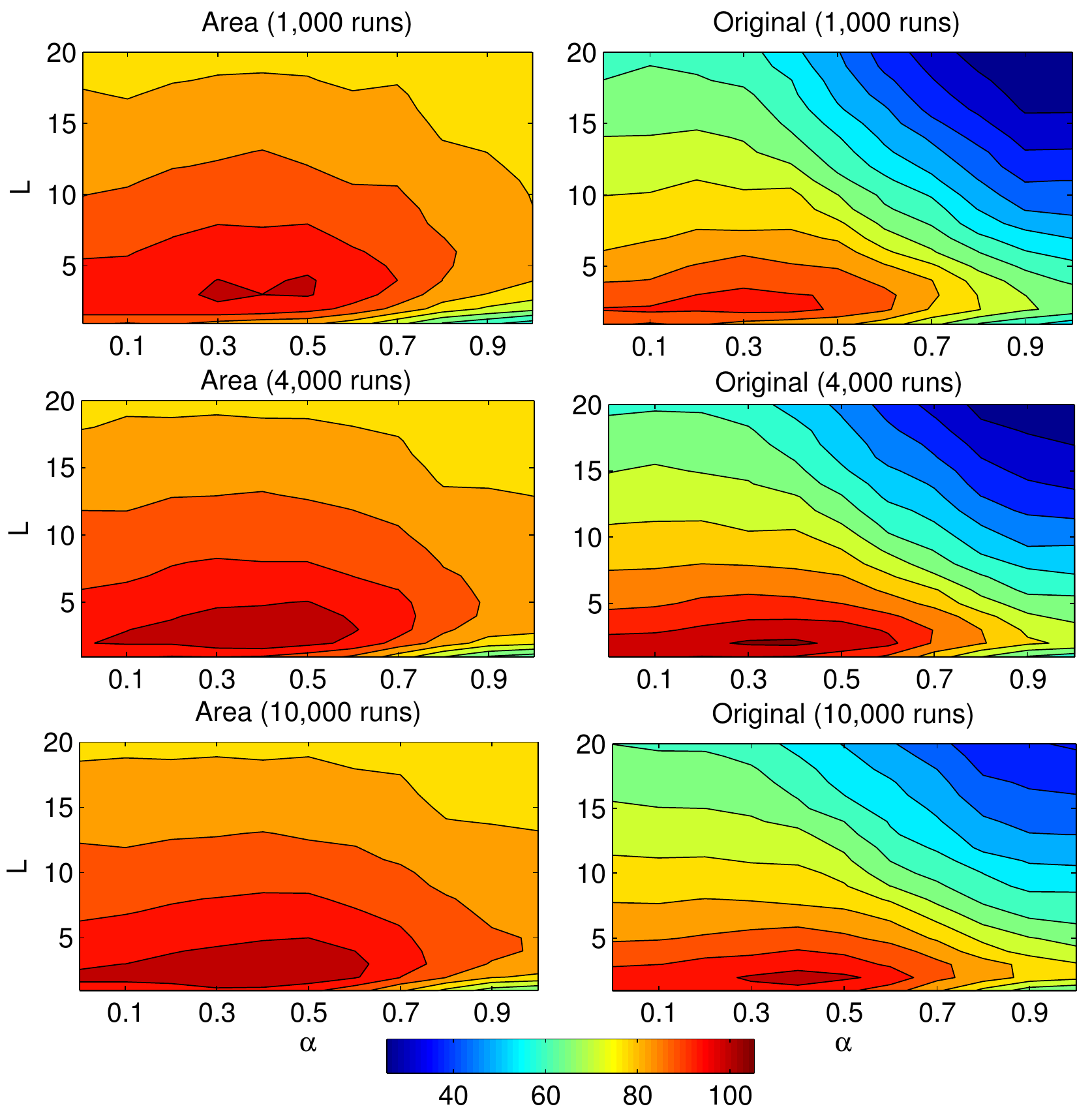}
\caption{Overall score for Network 1. From top to bottom, plots show the results for $R=1,000$, $R=4,000$ and $R=10,000$ for both the area (left) and the original (right) scoring settings, as a function of $\alpha$ and $L$.}
  \label{fig:HeatMapLalpha}
\end{figure}

Starting from the parameters used in Naive TIGRESS ($R=4,000$, $\alpha=0.2$ and $L=5$, original score), we assess the influence of the different parameters by systematically testing the following combinations:
\begin{itemize}
\item original (\ref{eq:originalscore}) or area (\ref{eq:areascore}) scoring method;
\item randomization parameter $\alpha\in\cbr{0,0.1 \ldots, 1}$;
\item length of the LARS path $L\in \cbr{1,2 \ldots 20}$;
\item number of randomization runs $R\in\cbr{1,000 ; 4,000 ; 10,000 }$.
\end{itemize}
Figure \ref{fig:HeatMapLalpha} summarizes the overall score \eqref{overall_score} obtained by each combination of parameters on Network $1$.

A first observation is that the \emph{area} scoring method consistently outperforms the \emph{original} scoring method, for any choice of $\alpha$ and $L$. This suggests that, by default, the newly proposed area score should be preferred to the classical original score. We also note that the performance of the area score is less sensitive to the value of $\alpha$ or $L$ than that of the original score. For example, any value of $\alpha$ between $0.2$ and $0.8$, and any $L$ less than $10$ leads to an overall score of at least $90$ for the area score, but it can go down to $60$ for the original score.
This is a second argument in favor of the \emph{area} scoring setting: as it is not very sensitive to the choice of the parameters, one may practically more easily tune it for optimal performance. On the contrary, the window of $(\alpha,L)$ values leading to the best performance is more narrow with the original scoring method, and therefore more difficult to find \emph{a priori}. The recommendation of \cite{Meinshausen2010Stability} to choose $\alpha$ in the range $[0.2,0.8]$ is clearly not precise enough for GRN inference. The best overall performance is obtained with $(\alpha=0.4,L=2)$ in both scoring settings.

\begin{figure}[!ht]
  \centering
  \includegraphics[width=\textwidth]{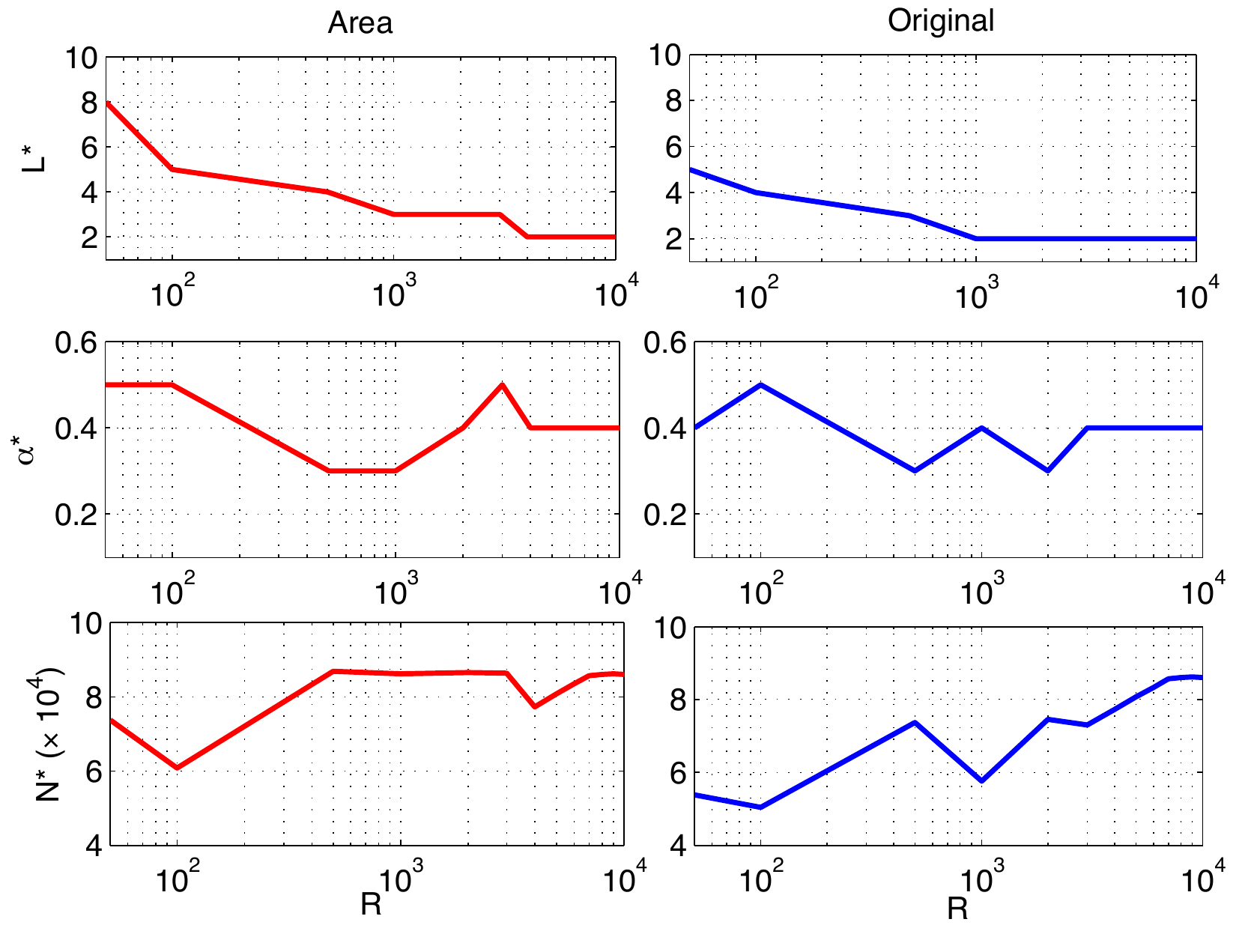}
\caption{Optimal values of parameters $L$, $\alpha$ and $N$ with respect to the number of resampling runs}
  \label{fig:bestparam}
\end{figure}

Regarding the relationship between $\alpha$ and $L$, we observe in Figure~\ref{fig:HeatMapLalpha} a slight positive correlation for the optimal $L$ as a function of $\alpha$, particularly for the area score. For example, for $R=10^4$, $L=2$ is optimal for $\alpha\leq 0.4$, but $L\geq 4$ is optimal for $\alpha \geq 0.8$. The effect is even more pronounced for $R=4,000$. This can be explained by the fact that when $\alpha$ increases, we decrease the variations in the the different runs of LARS and therefore reduce the diversity of features selected; increasing the number of LARS is a way to compensate this effect by increasing the number of features selected at each run. Another way to observe the need to ensure a sufficient diversity is to observe how the best parameters $L$ and $\alpha$ vary as a function of $R$ (Figure \ref{fig:bestparam}). It appears clearly that the optimal number of steps $L^*$ decreases when the number of resampling runs increases and stabilizes at $2$. This is not a surprising result. Indeed, when more resampling is performed, the chance of selecting a given feature increases. The number $N$ of non zero scores subsequently increases and it thus becomes unnecessary to look further in the regularization path. On the other hand, the value of $\alpha^*$ lies steadily between $0.3$ and $0.5$, suggesting that the adjustment to the number of bootstraps can mostly be made through the choice of $L$.

\begin{figure}[!ht]
  \centering
  \includegraphics[width=\textwidth]{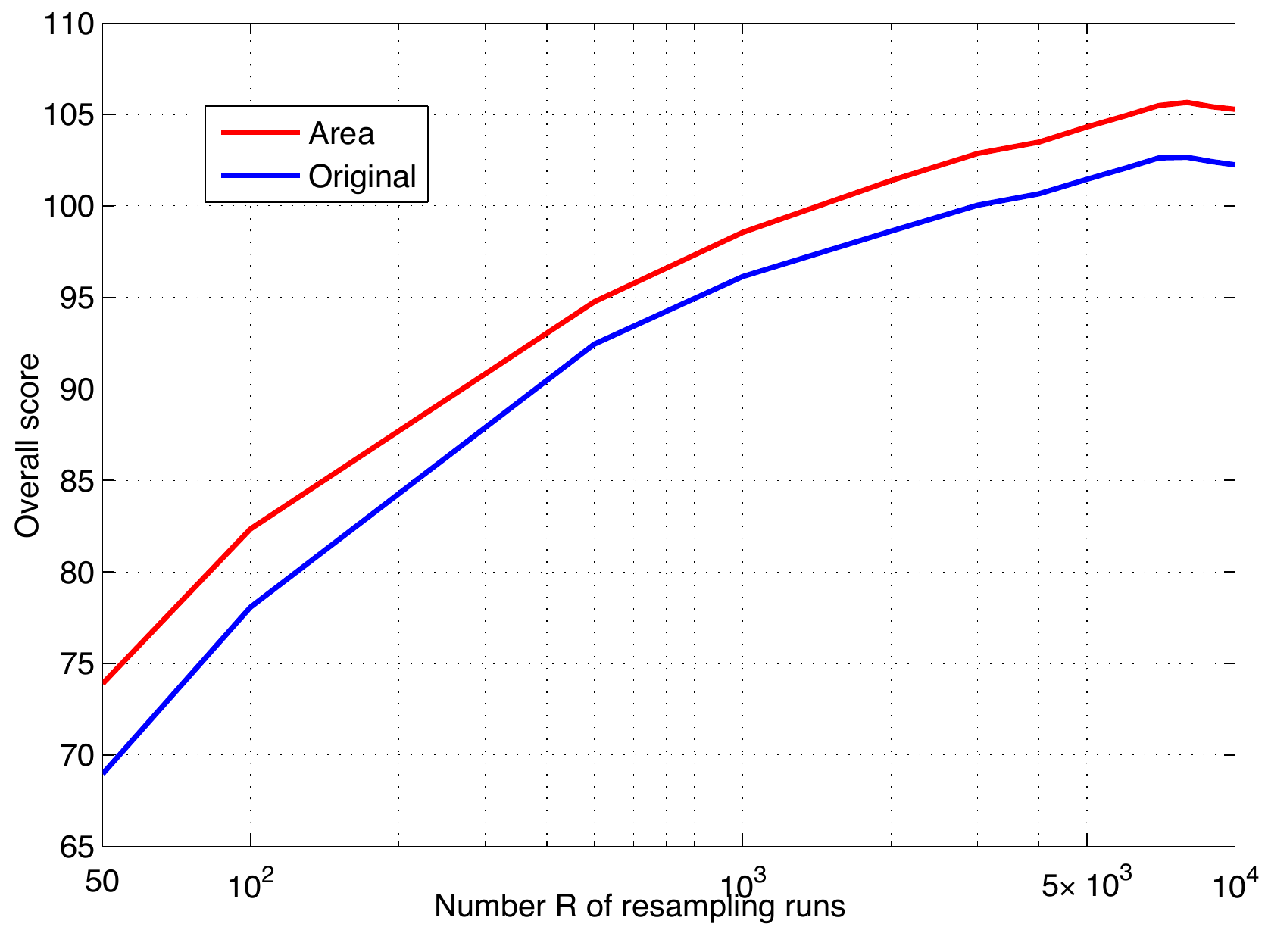}
\caption{Overall score as a function of $R$. In both scoring settings, $\alpha$ and $L$ were set to $0.4$ and $2$, respectively.}
  \label{fig:perfR}
\end{figure}

Finally, we unsurprisingly observe that increasing the number $R$ of resampling runs leads to better performances. On Figure \ref{fig:perfR}, we show the overall score as a function of $R$ with $L=2$ and $\alpha=0.4$. We clearly see that, for both scoring methods, increasing the number of runs is beneficial. The performance seems to reach an asymptote only when $R$ becomes larger than $5,000$. 
\OMIT{
The best performance is obtained for $R=8,000$ and reaches $105.68$, which beats all DREAM5 submissions for this network. This value corresponds to $AUPR=0.3197$ and $AUROC=0.7886$.
}

\subsection{Comparison with other methods}

\begin{figure}[!ht]
  \centering
  \includegraphics[width=\textwidth]{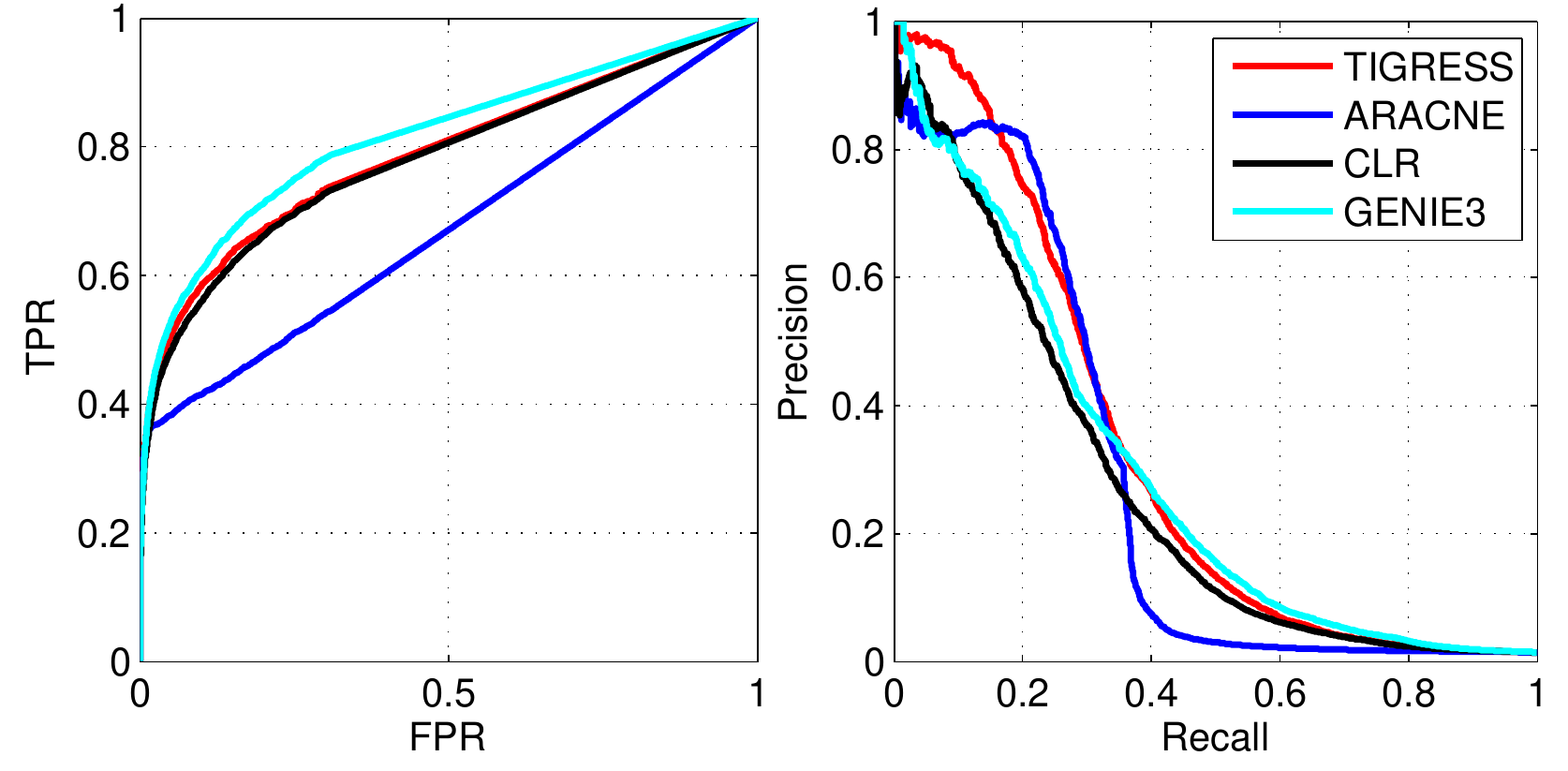}
\caption{ROC (left) and Precision/Recall (right) curves for several methods on Network $1$.}
  \label{fig:compare_net1}
\end{figure}

Figure \ref{fig:compare_net1} depicts both the ROC and the Precision/Recall curves for several methods on Network $1$. Table \ref{tab:compare_net1} summarizes these performances in terms of $AUPR$, $AUROC$ and related p-values as well as the overall score. Here, TIGRESS was run with $\alpha=0.4$, $L=2$ and $R=8,000$ which corresponds to the best performance of the algorithm, as investigated in the previous section.

\begin{table}[!ht] 
\centering 
\begin{tabular}{|l|c|c|c|c|c|} 
\hline 
\textbf{Method} & \textbf{AUPR} & \textbf{$p_{AUPR}$} & \textbf{AUROC} & \textbf{$p_{AUROC}$} & \textbf{Overall score}\\ \hline 
TIGRESS & 0.320 & 1.17e-145 & 0.789 & 3.74e-67 & 105.68 \\ \hline 
GENIE3 & 0.291 & 1.60e-104 & 0.815 & 3.06e-106 & 104.65 \\ \hline 
Naive TIGRESS & 0.301 & 7.20e-118 & 0.782 & 3.48e-59 & 87.80 \\ \hline
CLR & 0.265 & 1.82e-73 & 0.782 & 1.41e-58 & 65.30 \\ \hline 
ANOVA-based & 0.245 & 8.17e-53 & 0.780 & 1.34e-56 & 53.98 \\ \hline
ARACNE & 0.276 & 1.73e-85 & 0.672 & 9.82e-01 & 42.38 \\ \hline 
\end{tabular} 
\caption{Comparison of different methods on Network 1 of the DREAM5 challenge. The performance of GENIE3, Naive TIGRESS and ANOVA were obtained during the DREAM5 competition. TIGRESS corresponds to the choice of parameters leading to the best performance (area score, $\alpha=0.4$, $L=2$, $R=8,000$). We ran CLR and ARACNE using public implementations of these methods.} 
\label{tab:compare_net1} 
\end{table}

TIGRESS outperforms all methods in terms of AUPR and all methods but GENIE in terms of AUROC. Moreover, the shape of the Precision/Recall curve suggests that the top of the prediction list provided by TIGRESS contains more true edges than other methods. The ROC curve, on the other hand, focuses on the entire list of results. Therefore, we would argue that TIGRESS is more reliable than GENIE in its first predictions but contains overall more errors when we go further in the list.

\subsection{\emph{In vivo} networks results}
Since Naive TIGRESS did not perform very well on the \emph{in vivo} networks at the DREAM5 competition (Table~\ref{tab:dream_results}), we now test on these networks TIGRESS with the best parameters selected on the \emph{in silico} (area score, $\alpha=0.4$, $L=2$ and $R=10,000$). Table \ref{tab:res_nets34} shows the values of AUPR, AUROC, related p-values and overall score for DREAM5 networks $3$ and $4$ reached by TIGRESS.

\begin{table}[!ht] 
\centering 
\begin{tabular}{|l|c|c|c|c|c|} 
\hline 
\textbf{Network} & \textbf{AUPR} & \textbf{$p_{AUPR}$} & \textbf{AUROC} & \textbf{$p_{AUROC}$} & \textbf{Overall score}\\ \hline 
DREAM5 Network 3 & 0.0660 & 4.79e-06 & 0.5887 & 6.66e-02 & 3.25 \\ \hline 
DREAM5 Network 4 & 0.0199 & 5.86e-01 & 0.5143 & 2.02e-01 & 0.46 \\ \hline 
\end{tabular} 
\caption{TIGRESS performance on DREAM5 Networks 3 and 4.} 
\label{tab:res_nets34} 
\end{table}

\begin{figure}[!ht]
  \centering
  \includegraphics[width=.5\textwidth]{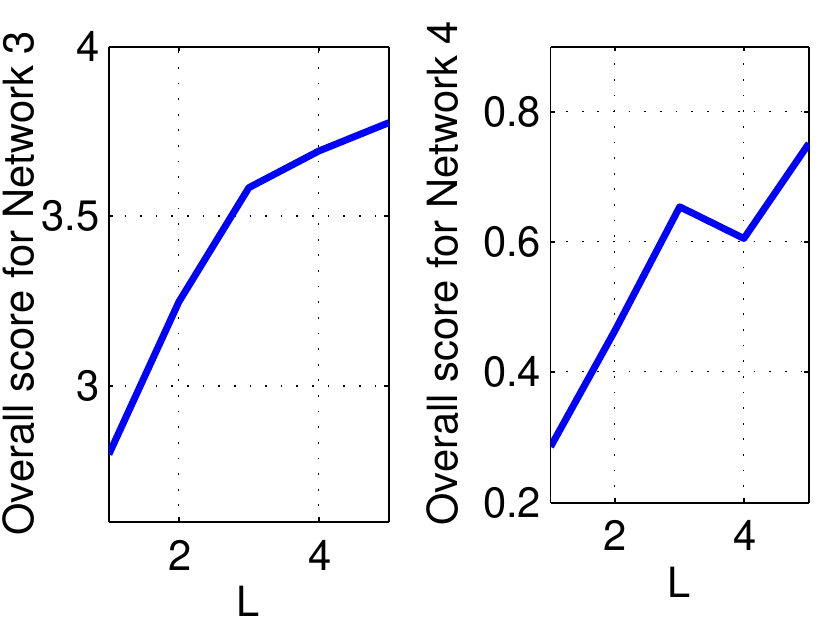}
\caption{Overall score with respect to $L$ for networks $3$ and $4$ ($\alpha=0.4$, $R=10,000$).}
  \label{fig:ovscoreL34}
\end{figure}

The results on these two networks are overall disappointing: TIGRESS does not do better than Naive TIGRESS. In fact, both sets of results are very weak. Without attempting to re-optimize all parameters for each network, one may wonder whether the parameters chosen using the \emph{in silico} network are optimal for the \emph{in vitro} networks. As a partial answer, Figure \ref{fig:ovscoreL34} shows the behavior of the overall score with respect to $L$ for Networks 3 and 4. Interestingly, it seems that a larger $L$ is preferable in this case, suggesting that one may have to adapt the parameters to the size of the network. Indeed, networks $3$ and $4$ contain respectively $4,511$ and $5,950$ nodes, making them much larger than the \emph{in silico} network we tuned the parameters on. However, the improvement is not dramatic in absolute value.

\begin{figure}[!ht]
  \centering
  \includegraphics[width=\textwidth]{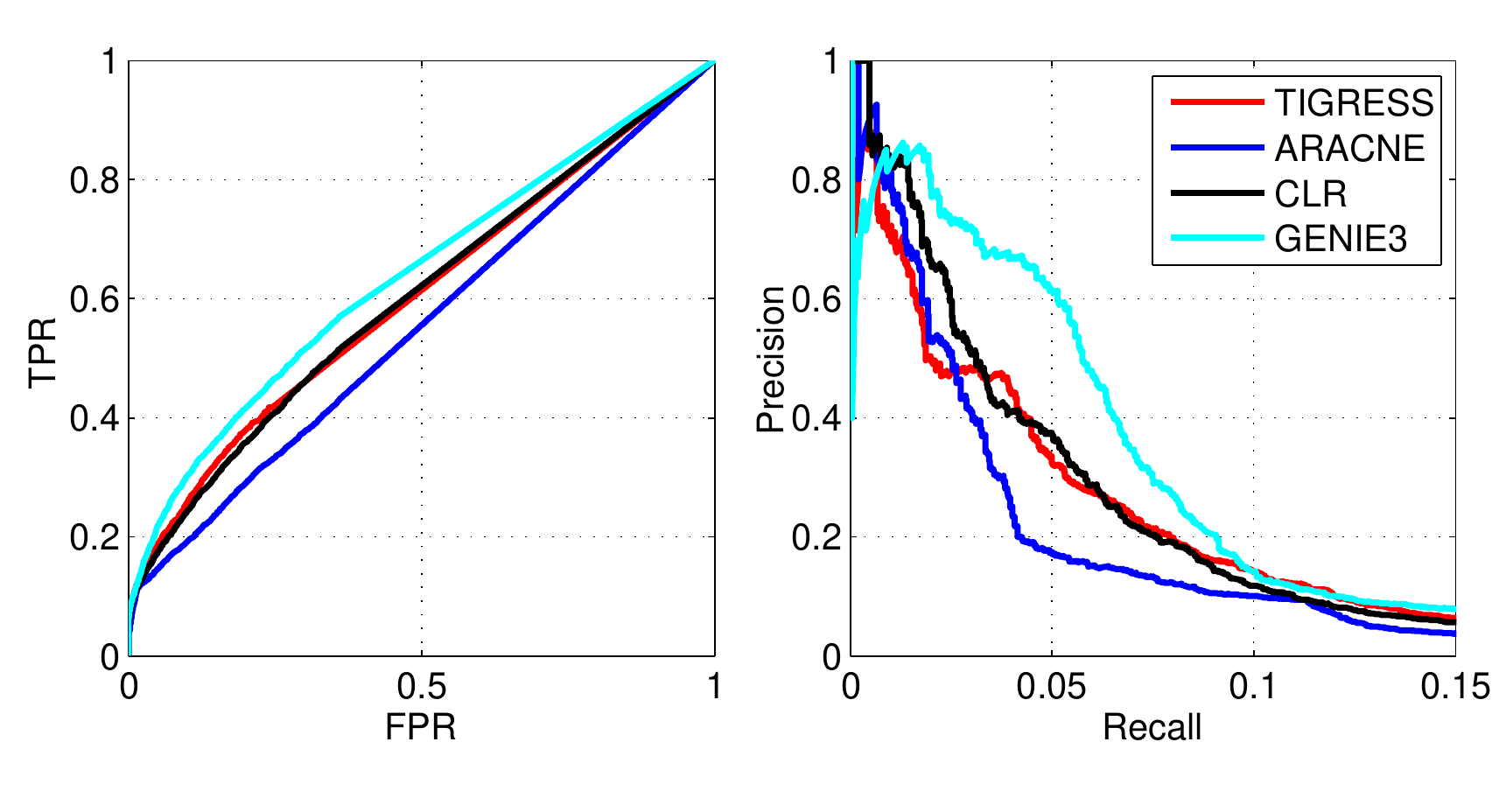}
\caption{Precision/Recall (Left) and ROC (Right) curves for several methods on the \ecoli{} dataset.}
  \label{fig:compare_ecoli}
\end{figure}

On Figure \ref{fig:compare_ecoli} we compare Precision/Recall and ROC curves obtained with TIGRESS with several other algorithms on the \ecoli{} network from \cite{Faith2007Large-scale}. Table \ref{tab:compare_ecoli} compares the areas under the curves. TIGRESS is comparable to CLR, while GENIE3 outperforms other methods. However the overall performance of all methods remains disappointing.

\begin{table}[!ht] 
\centering 
\begin{tabular}{|l|c|c|} 
\hline 
\textbf{Method} & \textbf{AUPR} & \textbf{AUROC}\\ \hline 
TIGRESS & 0.0624 & 0.6026 \\ \hline 
ARACNE & 0.0498 & 0.5531 \\ \hline 
CLR & 0.0641 & 0.6019 \\ \hline 
GENIE3 & 0.0814 & 0.6375 \\ \hline 
\end{tabular} 
\caption{TIGRESS compared to state-of-the-art methods on the \ecoli{} Network.} 
\label{tab:compare_ecoli} 
\end{table}

\subsection{Analysis of errors on \ecoli}

To understand further the advantages and limitations of TIGRESS, we analyse the type of errors it typically makes taking the \ecoli{} dataset as example. We analyse FP, \emph{i.e.} cases where TIGRESS predicts an interaction that does not appear in the gold standard GRN. 

\begin{figure}[!ht]
\centering
\includegraphics[width=.5\textwidth]{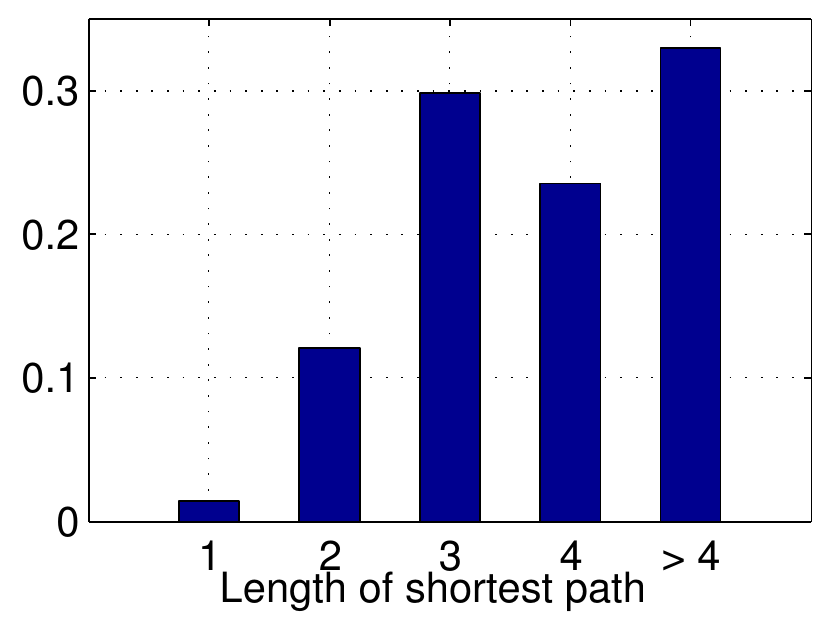}
\caption{Exact distribution of the shortest path between spuriously predicted TF-TG couples.}
\label{fig:exact_dist}
\end{figure}

We focus in particular on quantifying how far a wrongly predicted interaction is from a true one, and introduce for that purpose the notion of distance between two genes as the shortest path distance between them on the gold standard GRN, forgetting about the direction of edges. For two genes $G1$ and $G2$, we call  $G1$-$G2$ a \emph{distance-$x$} link if the shortest path between $G1$ and $G2$ on the true network has length $x$. Figure \ref{fig:exact_dist}, shows the distribution of these distances for spuriously discovered edges over the gold standard network, \emph{i.e.} the actual proportion of distance-$x$ links, with $x\in\{1,2,3,4,>4\}$. We write $\hat{p}_x$ the proportion of spurious TG-TF couples with distance $x$. 

\begin{figure}[!ht]
\centering
\includegraphics[width=\textwidth]{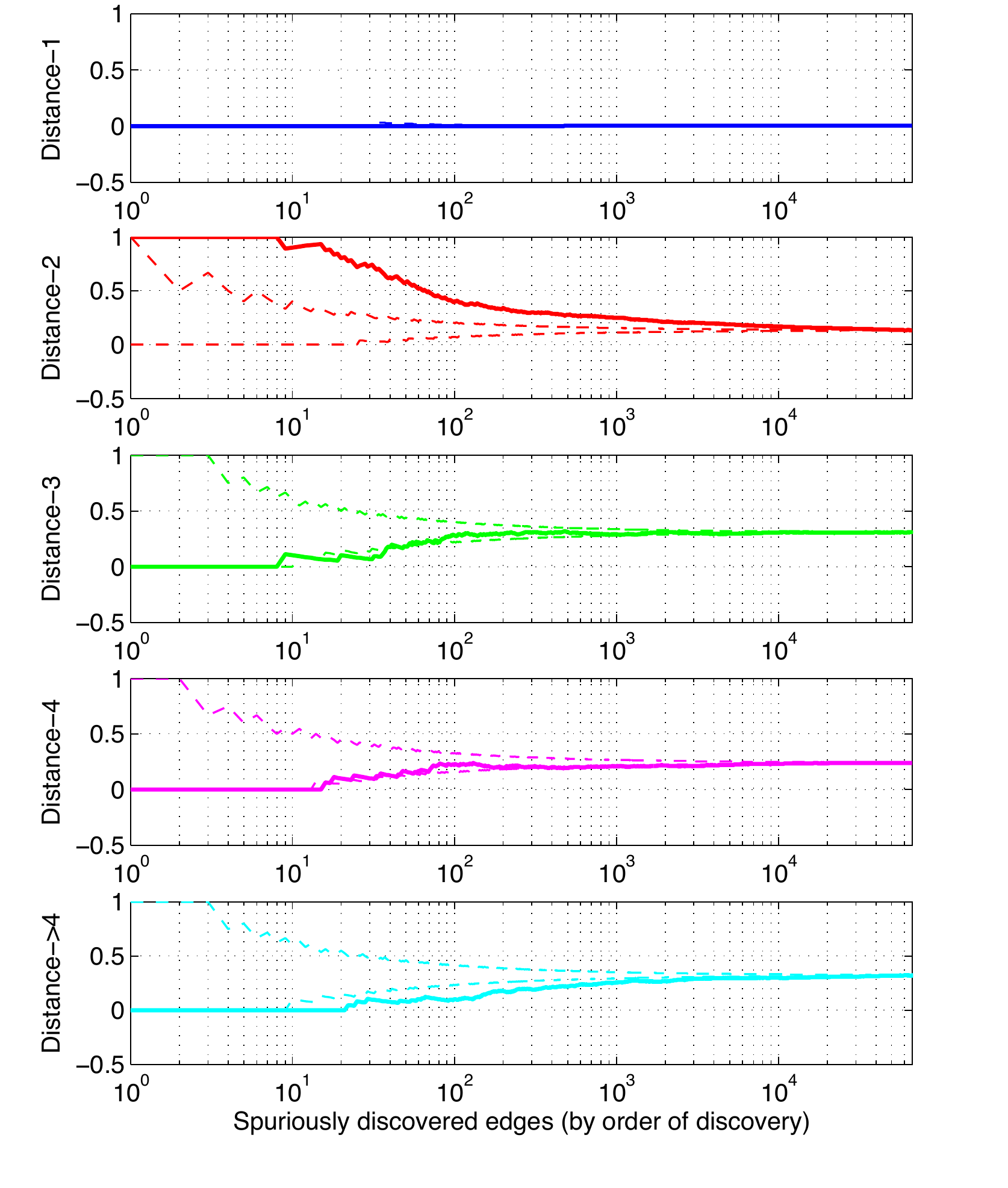}
\caption{Distribution of the shortest path length between nodes of spuriously detected edges and $95\%$ confidence interval for the null distribution. These edges are ranked by order of discovery.}
\label{fig:distshortestpath}
\end{figure}

Figure \ref{fig:distshortestpath} depicts the distribution of distance-$x$ proportions among the spuriously detected edges, as a function of the number of predicted edges. Dotted lines represent the $95\%$ confidence interval around the exact distribution $(\hat{p}_x)_x$. For a given number $r$ of spuriously predicted edges, this interval is computed as 
$$\left[\frac{q_{0.025}(\hat{p}_x)}{r} ; \frac{q_{0.975}(\hat{p}_x)}{r} \right],$$ 
where $q_{a}(\hat{p}_x)$ represents the quantile of order $a$ of a hypergeometric distribution $\mathcal{H}(N_S,\hat{p}_xN_S,r)$ and $N_S$ is the total number of spuriously predicted edges. 

We observe that most of the recovered false positives appear as distance-$2$ edges in a significantly higher proportion than $\hat{p}_2$ whereas significantly less distance-$>4$ edges are discovered. These results strongly suggest that most of TIGRESS errors - especially at the top of the list - are indeed sensible guesses, where the two nodes, spuriously discovered with a parent/child relationship are actually separated by only one other node. In Table \ref{tab:family}, we detail the three possible patterns observable in this situation.

\begin{table}[!ht]
\centering
\begin{tabular}{|c|c|p{5cm}|}
\hline
Name & Illustration & Description\\ \hline \hline
Siblings &  \parbox[c]{.2\textwidth}{\includegraphics[width=.2\textwidth]{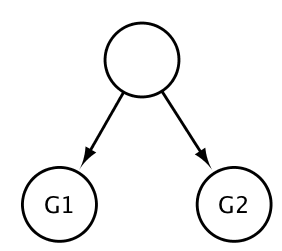}} &  G1 and G2 have a common parent. They are \emph{siblings}.\\ \hline
 Couple &  \parbox[c]{.2\textwidth}{\includegraphics[width=.2\textwidth]{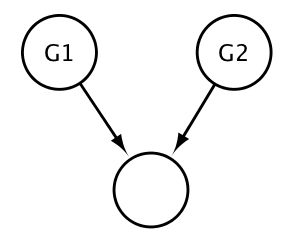}} & G1 and G2 have a common child. They are a \emph{couple}. \\  \hline
Grandparent/Grandchild &  \parbox[c]{.2\textwidth}{\includegraphics[width=.2\textwidth]{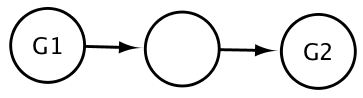}}  & G1 has a child that is a parent of G2. G1 is a \emph{grandparent} of G2. \\ \hline \hline
\end{tabular}
\caption{Distance-$2$ patterns between two nodes G1 and G2 in a directed graph.}
\label{tab:family}
\end{table}

\begin{figure}[!ht]
\centering
\includegraphics[width=\textwidth]{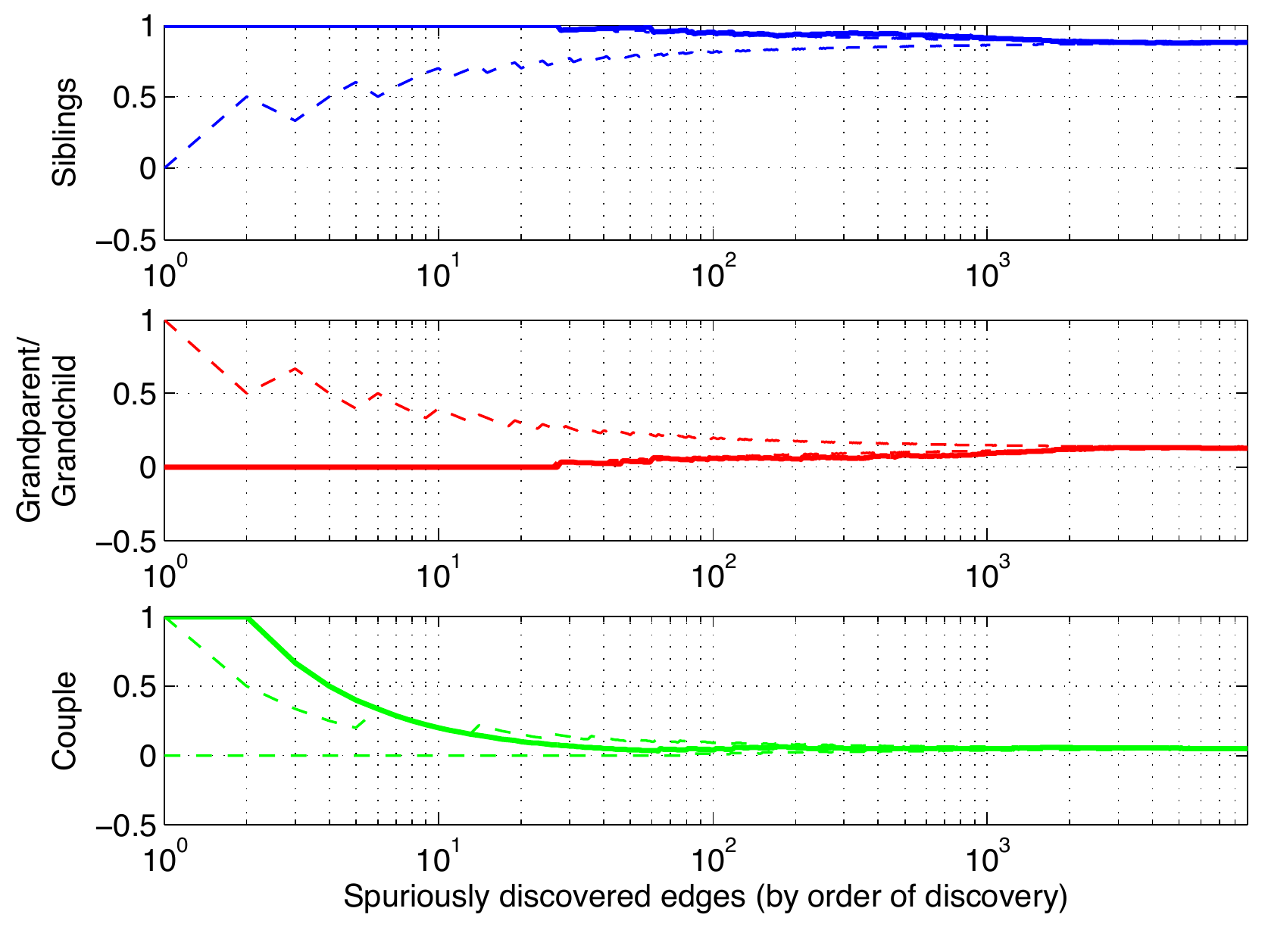}
\caption{Distribution of distance $2$ errors. $95\%$ error bars were computed using the quantiles of a hypergeometric distribution.}
\label{fig:dist2errors}
\end{figure}

Figure \ref{fig:dist2errors} focuses on distance-$2$ errors. Note that some edges show more than one pattern, \emph{e.g.} the first spurious edges are both \emph{siblings} and \emph{couples}. It appears that most of them are \emph{siblings} and can thus be interpreted as spurious feed-forward loops. We believe that this can be explained by three main reasons: i) the discovered edges actually exist but have not been experimentally validated yet; ii) there is more of a linear relationship between siblings than between parent and child; iii) some nodes have very correlated expression levels, making it difficult for TIGRESS to tell between the parent and the child.

\section{Discussion}

In this paper, we presented TIGRESS, a new method for GRN inference. TIGRESS expresses the GRN inference problem as a feature selection problem, and solves it with the popular LARS feature selection method combined with stability selection. It ranked in the top 3 GRN inference methods at the 2010 DREAM5 challenge, without any parameter tuning. We clarified in this paper the influence of each parameter, and  showed that further improvement may result from finer parameter tuning.

We proposed in particular a new scoring method for stability selection, based on the area under the stability curve. It differs from the original formulation of \cite{Meinshausen2010Stability} which does not take into account the full distribution of ranks of a feature in the randomized feature selection procedure. Comparing the two, we observed that the new area scoring technique yields better results and is less sensitive to the values of the parameters: practically all values of, \emph{e.g.}, the randomization parameter $\alpha$ yield the same performance. Similarly, the choice of the number $L$ of LARS steps to run seems to have much less impact on the performance in this new setting. \OMIT{We observed a drift of the optimal number of LARS steps with the increase of the number of subsampling runs: the more subsampling runs, the smaller the optimal value of $L$. This can easily be explained. Indeed, as more runs are performed, more TFs are considered, making it unnecessary to look further in the regularization path.} As we showed, the original and area scores for a feature $t$ can be both expressed in a common formalism as $E\left[\phi(H)\right]$ for different functions $\phi$, where $H_t$ is the rank of feature $t$ as selected by the randomized LARS. It could be interesting to systematically investigate variants of these scores with more general non-increasing functions $\phi$, not only for GRN inference but also more generally as a generic feature selection procedure.

Comparing TIGRESS - as tuned optimally - to state-of-the-art algorithms on the \emph{in silico} network, we observed that it achieves a similar performance to that of GENIE3 \citep{Huynh-Thu2010Inferring}, the best performer at the DREAM5 challenge. However, TIGRESS does not do as good as this algorithm on \emph{in vivo} networks. GENIE3 is also an ensemble algorithm but differs from TIGRESS in that it uses a non-linear  tree-based method for feature selection, while TIGRESS uses LARS. The difference in performance could be explained by the fact that the linear relationship between TGs and TFs assumed by TIGRESS is far-fetched given the obvious complexity of the problem.

A further analysis of our results on the \ecoli{} network from \cite{Faith2007Large-scale} showed that many spuriously detected edges follow the same pattern: TIGRESS discovers edges when in reality the two nodes are \emph{siblings}, and thus tends to wrongly predict feed-forward loops. This result suggests many directions for future work. Among them, we believe that operons, \emph{i.e.} groups of TGs regulated together could be part of the problem. Moreover, it could be that there is more of a linear relationship between siblings than between parent and child, as TFs are known to be operating as \emph{switches}, \emph{i.e.} it is only after a certain amount change in expression of the TF that related TGs are affected. However, it is worth noting that \emph{in vivo} networks gold standards may not be complete. Therefore, the hypothesis that TIGRESS is actually correct when predicting these loops cannot be discarded. 

While it seems indeed more realistic not to restrict underlying models to linear ones, it is fair to say that no method performs very well in absolute values on \emph{in vivo} networks. For example, performances on the \ecoli{} network seem to level out at some $64\%$ AUROC and $8\%$ AUPR which cannot be considered satisfying. This suggests that while regression-based procedures such as TIGRESS or GENIE3 are state-of-the-art for GRN inference, their performances seem to hit a limit which probably cannot be outdistanced without some changes in the global approach such as adding some supervision in the learning process as, {\textit e.g.}, investigated in \citep{Mordelet2008SIRENE}.

\section{Acknowledgements}
JPV was supported by ANR grant ANR-09-BLAN-0051-04 and ERC grant SMAC-ERC-280032.

\bibliographystyle{plain}
\bibliography{tigressTR}

\end{document}